\documentclass[letterpaper, 10 pt, conference]{ieeeconf}  

\IEEEoverridecommandlockouts                              

\usepackage[dvipsnames]{xcolor}
\usepackage{graphics} 
\usepackage{epsfig} 
\usepackage{caption}
\usepackage{subcaption}
\usepackage{graphicx}
\usepackage{booktabs}
\let\labelindent\relax
\usepackage{nicematrix,enumitem}
\usepackage{hyperref}  
\usepackage{amssymb}
\usepackage{algorithm,algorithmic}
\usepackage{multirow}

\hypersetup{
    colorlinks=true,
    linkcolor=blue,
    filecolor=magenta,      
    urlcolor=cyan,
    pdftitle={Overleaf Example},
    pdfpagemode=FullScreen,
    }

\title{\LARGE \bf
Informed along the road: roadway capacity driven graph convolution network for network-wide traffic prediction}

\author{Zilin Bian$^{1}$\textsuperscript{,*}, Jingqin Gao$^1$, Kaan Ozbay$^{1}$, Fan Zuo$^1$, Dachuan Zuo$^1$, Zhenning Li$^{2}$
\thanks{$^1$ Tandon School of Engineering, New York University, NY, USA, 10012. }%
\thanks{$^2$ State Key Laboratory of Internet of Things for Smart City, University of Macau, N21 5025a, Avenida da Universidade, Taipa, Macau, People's Republic of China}
\thanks{* Corresponding Authors email: {\tt\small zb536@nyu.edu}}%
}

\begin{document}

\maketitle
\thispagestyle{empty}
\pagestyle{empty}
\begin{abstract}
    
While deep learning has shown success in predicting traffic states, most methods treat it as a general prediction task without considering transportation aspects. Recently, graph neural networks have proven effective for this task, but few incorporate external factors that impact roadway capacity and traffic flow. This study introduces the Roadway Capacity Driven Graph Convolution Network (RCDGCN) model, which incorporates static and dynamic roadway capacity attributes in spatio-temporal settings to predict network-wide traffic states. The model was evaluated on two real-world datasets with different transportation factors: the ICM-495 highway network and an urban network in Manhattan, New York City. Results show RCDGCN outperformed baseline methods in forecasting accuracy. Analyses, including ablation experiments, weight analysis, and case studies, investigated the effect of capacity-related factors. The study demonstrates the potential of using RCDGCN for transportation system management.

\hfill\break%
\noindent\textit{Keywords}: Traffic prediction, Graph convolution, Roadway capacity, Traffic incident
\end{abstract}
\section{Introduction}

Traffic state prediction is crucial for efficient transportation system management, including both operational and planning aspects. However, predicting network-wide traffic states remains challenging due to inherent uncertainties such as traffic accidents and adverse weather conditions. Historically, research in this field has utilized model-driven approaches, like traditional statistical models \cite{kumar2015short}, to describe traffic flow properties, or data-driven methods, which leverage machine learning algorithms to harness large-scale real-time traffic data \cite{huang2014deep}. As artificial intelligence has evolved, early deep learning applications drew from time-series model concepts, similar to those used in natural language processing and stock price forecasting, to predict traffic states \cite{selvin2017stock}.

These initial efforts predominantly used recurrent neural network (RNN) variants, such as Long Short-Term Memory (LSTM) and Gated Recurrent Units (GRU) \cite{cho2014properties}, to model traffic flow. Researchers also explored convolutional neural network (CNN) variants \cite{ma2017learning}, although these models represent traffic data using grid-shaped structures in Euclidean space, which do not accurately reflect the non-Euclidean structure of real transportation networks.

Graph Convolutional Networks (GCNs) and their variants have seen significant success in addressing this limitation by projecting traffic data into non-Euclidean space and using graph structures to represent transportation networks, where nodes and edges represent road segments and their connectivity, respectively. While GCNs typically employ either a fixed adjacency matrix or an attention mechanism to capture spatial dependencies among road segments, each method has its drawbacks. Fixed adjacency matrices assume uniform influence among connected roads, whereas attention mechanisms determine traffic influence solely based on the state of neighboring segments, neglecting other potentially critical factors like roadway capacity.

According to the Highway Capacity Manual \cite{manual2010hcm2010}, roadway capacity is defined as the maximum traffic volume a road segment can handle under prevailing conditions, which crucially influences traffic flow. Current GCN models often overlook the varying capacity of roadways and other dynamic factors such as traffic incidents that can alter capacity. While some studies have incorporated external factors like weather or points of interest using temporal models, these do not directly affect roadway capacity or the traffic propagation process \cite{liao2018deep,bian2021time}.

This study introduces a novel Roadway Capacity-driven Graph Convolution Network (RCDGCN) model that integrates both static and dynamic roadway capacity factors to more accurately predict traffic states. Our contributions are:

\begin{itemize}
\item Development of the RCDGCN model, incorporating heterogeneous capacity-related factors from multiple data sources to predict traffic states.
\item Design of a Traffic Attention Block (TAB) within the RCDGCN, inspired by the Multinomial Logit Model, which adjusts traffic propagation intensities based on roadway capacity factors.
\item Validation of the model using real-world datasets from the ICM-495 highway and urban road networks in Manhattan, New York City, demonstrating superior performance over existing models.
\item An ablation study highlighting the importance of external capacity factors in improving prediction accuracy.
\item Enhanced model interpretability through analysis of learned weight parameters in TAB, identifying critical roadway links in the transportation network.
\end{itemize}
\section{Related Works}
In mode choice modeling, utility-based models assume travelers select the alternative with the maximum utility derived from factors of available alternatives. The utility function $U_i$ for alternative $i$ in the choice set $C$ is:
\begin{align}
U_i = V_i + \epsilon_i \label{eq:1}
\end{align}
Where $V_i$ is the estimated utility proportion, and $\epsilon_i$ is the unknown utility portion. Alternative $i$ is subject to constraints, and its utility function is detailed as:
\begin{align}
V_i = \beta_1+\beta_2\times S_2 + ... + \beta_l \times S_l\label{eq:2}
\end{align}
Where $S$ denotes variables for mode choice $i$, $l$ is the number of variables, and $\beta$ represents variable coefficients.
The multinomial logit model (MNL) is a utility-based approach that assumes independence between alternatives' variables in the utility function. MNL evaluates the utility of individual alternatives using probabilities:
\begin{align}
P_i = \frac{exp(V_i)}{\sum_{i=1}^{j} exp(V_i)} \label{eq:3}
\end{align}
Where $P_i$ is the probability of choosing alternative $i$ out of $j$ alternatives. The model choice converts to selecting the utility with the highest probability.
Similar to travelers' mode choice (walk, car, transit), the traffic propagation process along a road segment faces a mode choice problem at a macro level. The utility of mode choice is affected by available alternatives' attributes/characteristics. Likewise, traffic propagation from a road segment to its neighbors is affected by the road segments' attributes/characteristics.
Specifically, traffic propagation from a road segment mainly depends on the roadway capacity, indicated by capacity-related factors such as road geometry and traffic incidents. Representing these capacity-related factors as alternatives, the traffic propagation intensity can be calculated using the MNL utility functions.

\section{Methodology}

In this section, we introduce a novel GCN-based model that can integrate both the static and dynamic roadway capacity factors into the modeling of traffic propagation and prediction of traffic states. \par 

\subsection{Notations and problem definition}
\paragraph{Definition 1} Road network $\mathcal{G}$. Denote a general transportation network as a directed graph $\mathcal{G = (V,E)}$. In terms of the representation in the transportation network, $\mathcal{V}$ represents traffic sensors on the road segments and $\mathcal{E}$ represents the connectivity between traffic sensors. Denote $A \in \mathbb{R}^{N\times N}$ as the adjacency matrix in $\mathcal{G}$, representing the topology of road segments in the transportation network.

\paragraph{Definition 2} Traffic state matrix $X$. Denote the traffic states at time step $t$ as $X_t \in \mathbb{R}^{N\times P}$, where $P$ represents the dimension of traffic states such as traffic volume, traffic speed or travel time of each node in $\mathcal{G}$. 
\paragraph{Definition 3} Capacity feature matrix $Z$. Denote capacity-related features that affect the traffic propagation process along the road at time $t$ as $Z_t \in \mathbb{R}^{N\times L}$, where $L$ represents there are $L$ types of capacity-related factors included in $Z$.\par 

The problem of this study is to learn a mapping function $F(\cdot)$ given the observations of traffic states at $N$ nodes of historical $Q$ time steps. The historical observations of traffic states are denoted as $X = (X_{t_1}, X_{t_2}, ..., X_{t_Q})\in \mathbb{R}^{Q\times N \times P}$. The mapping function $F(\cdot)$ is represented as follows:
\begin{align}
    \hat{X} = F(X, Z) \label{eq:8}
\end{align}

In this study, the mapping function $F(\cdot)$ is used to predict next T time steps of traffic states $\hat{X} = (\hat{X}_{t_{Q+1}}, \hat{X}_{t_{Q+2}}, ..., \hat{X}_{t_{Q+T}})$, given $Q$ previous traffic states $X$ and external features $Z$.

\subsection{Roadway Capacity factors that affect the traffic propagation}
The traffic propagation process along the road segments is mainly determined by the roadway capacity. In this study, we will focus on analyzing external factors that can be used to indicate the roadway capacity in both static and dynamic perspectives.\par 
\textbf{Static capacity factor:} In this study, we adopted features such as roadway type, road segment length, total number of lanes and annual average daily traffic (AADT) to represent the static roadway capacity.\par 
\textbf{Dynamic capacity factor:} Traffic incidents (e.g., accidents, temporal work zones) often cause temporary lane closure and lead to dynamic capacity loss. In this study, we introduced dynamic capacity factors the occurrence of a traffic incident and the open lane ratio resulted by the incident. The open lane ratio is calculated by the comparison between the number of lanes closed/blocked and the total number of lanes on the road. \par

\subsection{Framework of roadway capacity driven graph convolution networks}
\begin{figure}[htp]
    \centering
    \includegraphics[width=1\columnwidth]{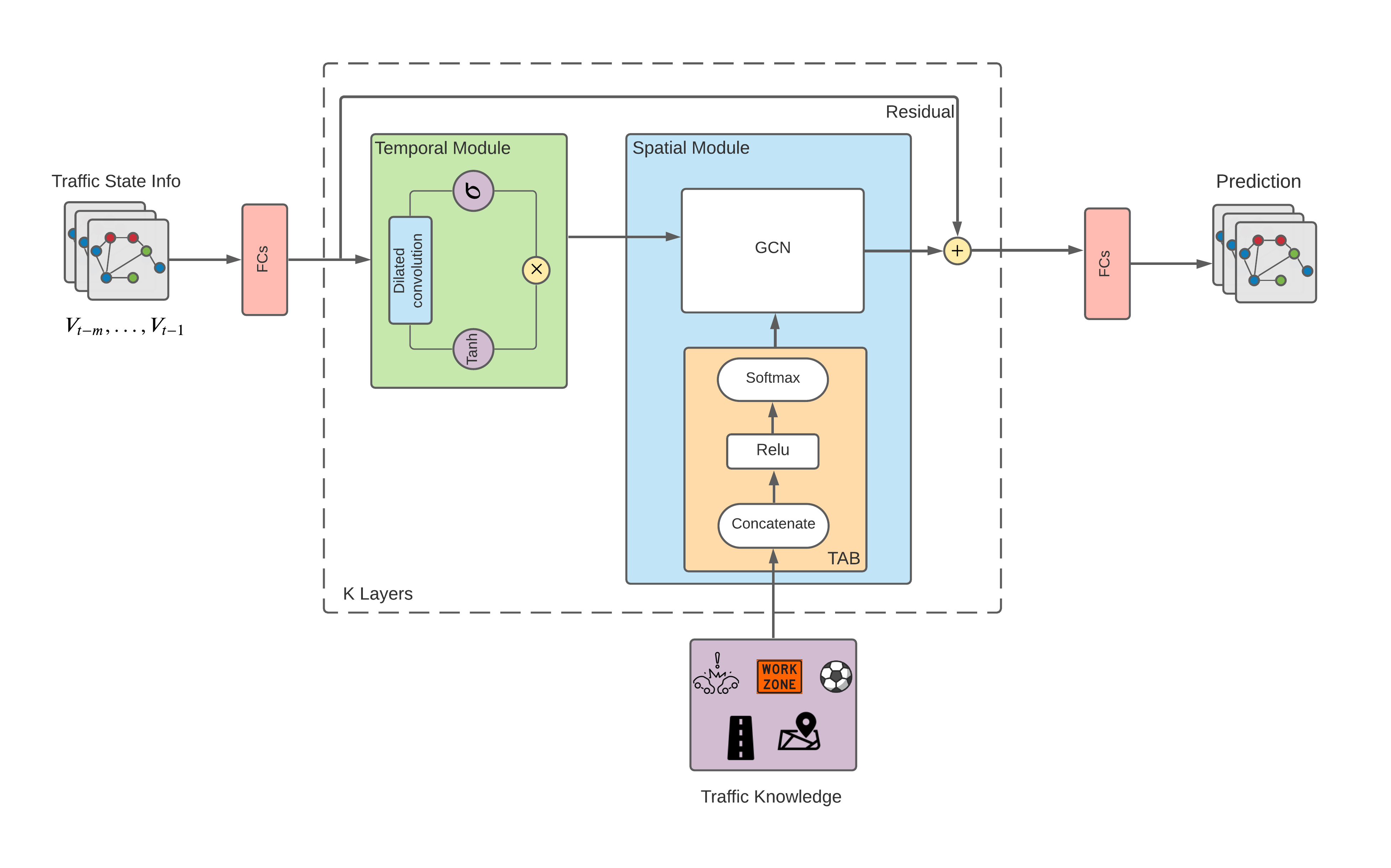}
    \caption{Framework of roadway capacity driven graph convolution network (RCDGCN)}
    \label{fig:RCDGCNStructure}
\end{figure}

The proposed model consists of the spatial and temporal modules, respectively. The spatial module introduces a newly designed traffic attention block (TAB) which can account for multiple capacity-related factors to model the traffic propagation using similar mechanism as MNL. The TAB learns the effect of each capacity-related factors and creates a traffic attention matrix which serves as the adjacency matrix for the model. The temporal module adopts temporal convolution network (TCN) to learn the dependencies in traffic states along the temporal dimensions. The framework of the designed model (RCDGCN) is shown in Fig.\ref{fig:RCDGCNStructure}.\par 

\subsection{Temporal Convolution Network}
This study utilizes a Temporal Convolution Network (TCN) to model the temporal dependencies in traffic state data, as proposed by \cite{dauphin2017language}. Unlike Recurrent Neural Network (RNN) models, which may suffer from gradient explosion due to recursive propagation, TCN avoids this issue by using causal dilated convolutions. We have enhanced the TCN with two gating units, tanh and sigmoid, to control the flow through the network layers effectively:

\begin{align}
    h = \tanh(W_1 * x) \odot \sigma(W_2 * x) \label{eq:9}
\end{align}

The causal dilated convolution increases the receptive field exponentially without requiring a large parameter set, enabling the model to capture both local and global context efficiently:

\begin{align}
    \mathbf{x} * \mathbf{f} (t) = \sum_{i=0}^K \mathbf{f}(i) \mathbf{x}(t-d\times i) \label{eq:10}
\end{align}

\subsection{Traffic Attention Block}
The Traffic Attention Block (TAB), inspired by the multinomial logit model (MNL), is designed to incorporate and mimic the influence of both static and dynamic roadway capacity factors on traffic propagation. TAB adapts the utility-based approach of MNL to the network context by defining the utility between nodes based on travel time and external capacity factors:

\begin{align}
    V_{ij} = \beta_{ij}^0 \times TT_{ij} + \sum_{1}^L (\beta_{ij}^L \times Z_{ij}^L) \label{eq:11}
\end{align}

Using embedded features, TAB transforms the utilities into intensity scores for traffic propagation, which are then normalized using a softmax function:

\begin{multline}
    e_{ij} = W_{i}^0 \times X_i + \sum_{1}^L (W_{i}^L \times Z_{i}^L) \\
    + W_{j}^0 \times X_j + \sum_{1}^L (W_{j}^L \times Z_{j}^L) \label{eq:12}
\end{multline}

\begin{align}
    \alpha_{ij} = softmax(e_{ij}) = \frac{\exp(e_{ij})}{\sum_{1}^p exp(e_{ip})} \label{eq:13}
\end{align}

This process results in a traffic attention matrix $\Tilde{A}$, which is utilized in the graph convolution process to compute spatial dependencies effectively.

\subsection{Graph Convolution with Traffic Attention Matrix}
Utilizing the constructed traffic attention matrix $\Tilde{A}$, our graph convolution layer aggregates traffic state information across multiple neighborhood hops, enabling dynamic spatial dependency modeling:

\begin{align}
    H = \sigma(\sum_{\gamma=0}^\Gamma \Tilde{A}_\gamma X W_\gamma) \label{eq:15}
\end{align}

The output of this layer predicts future traffic states, optimizing the model using a mean square error loss function, thus providing a robust framework for real-time traffic state prediction across varied network conditions. This integrated approach not only improves prediction accuracy but also enhances the model's ability to adapt to dynamic road conditions influenced by various capacity factors.

\section{Experiments}
The proposed Roadway Capacity-driven Graph Convolution Network (RCDGCN) integrates various capacity-related factors and was evaluated using two distinct datasets: the ICM-495 dataset and the Manhattan dataset.\par

\textbf{ICM-495 dataset:} Provided by NYCDOT, this dataset includes traffic speed data aggregated at 5-minute intervals from January 1 to February 28, 2018, covering 393 segments of the ICM-495 highway system. It includes a range of road types, such as highways and urban streets. Traffic incident data, including time and lane closures, were combined with traffic speeds using spatial and temporal matching. Key features are traffic speed $X$, incident occurrence $I$, and open lane ratio $O$.\par

\textbf{Manhattan dataset:} Sourced from NPMRDS, this dataset features probe vehicle traffic speeds from the complex network of 632 segments in Midtown and Downtown Manhattan for the year 2019. Roadway geometry data (e.g., road type and number of lanes) was added to estimate maximum roadway throughput. Input features include traffic speed $X$, road segment type $L$, and maximum throughput $MT$.\par

Both datasets were normalized to the range [0,1] for model training and testing. Table \ref{tab:external} summarizes the capacity factors considered: incident occurrence ($I$) and open lane ratio ($O$) for ICM-495, and road type ($L$) and maximum throughput ($MT$) for Manhattan.\par 

\begin{figure}[!tbp]
  \centering
  \subfloat[]{\includegraphics[width=0.633\columnwidth]{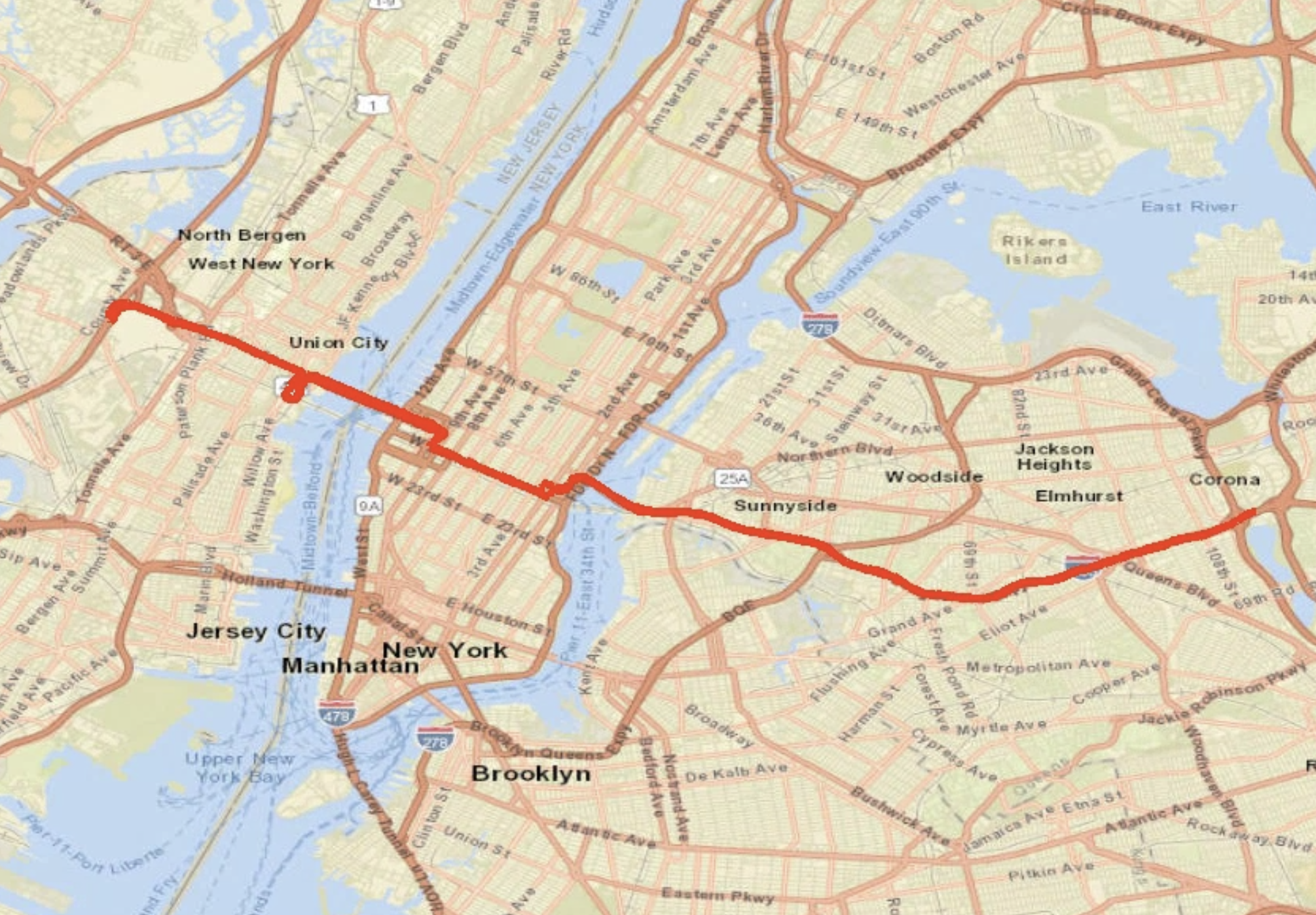}\label{fig:495network}}
  \hfill
  \subfloat[]{\includegraphics[width=0.362\columnwidth]{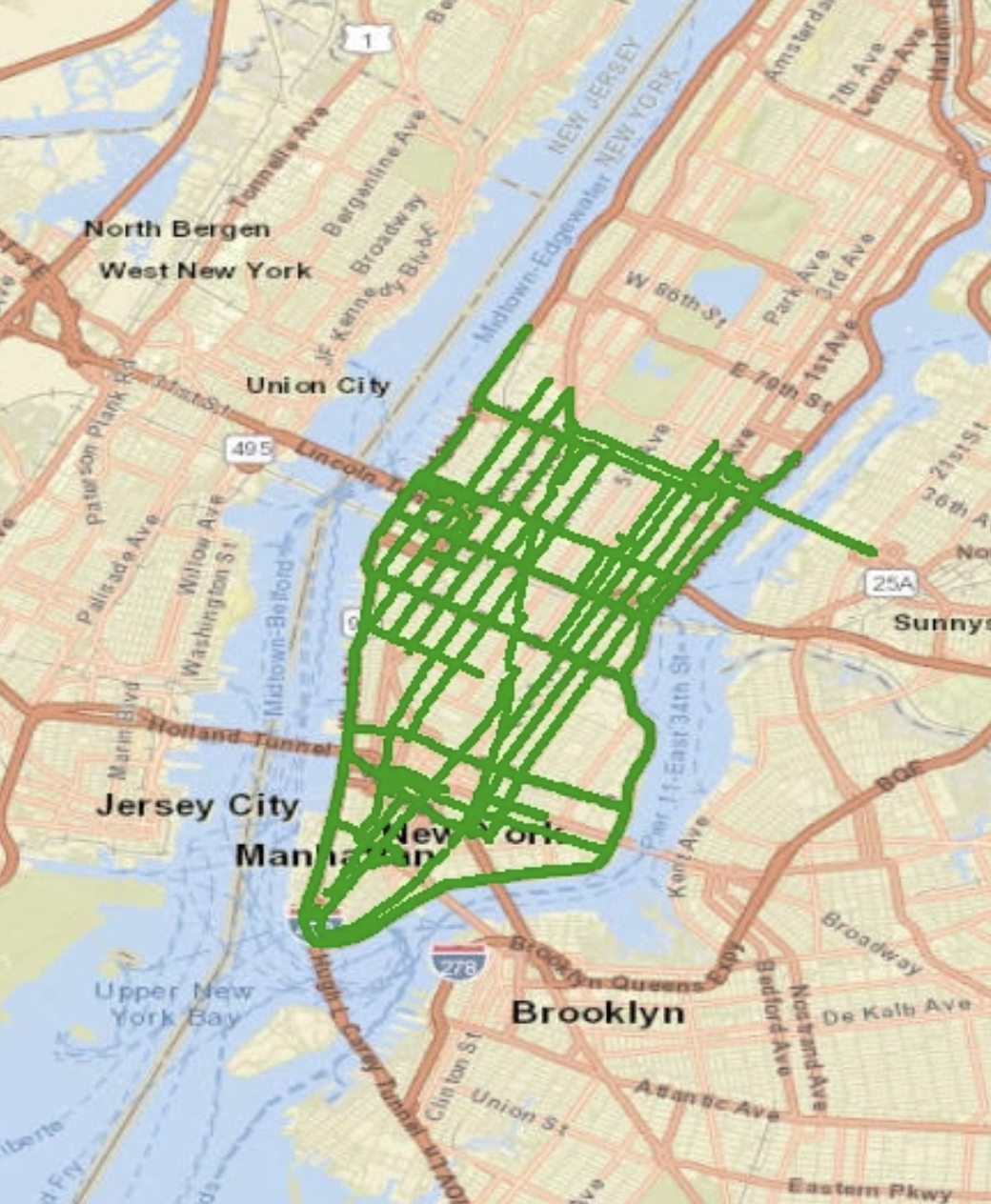}\label{fig:middownnetwork}}
  \caption{Network layout of ICM-495 (a), Manhattan (b)}
  \label{fig:twonetwork}
\end{figure}

\begin{table}[!ht]
    \caption{Data description of capacity factors}
    \label{tab:external}
    \centering
    \small
    \setlength{\tabcolsep}{3pt} 
    \begin{tabular}{l l c c}
        \hline
        Capacity factor & Description & Notation & Dataset \\
        \hline
        Occurrence of \\an incident & Incident presence & $I$ & ICM-495 \\
        Open lane ratio & Open lanes count & $O$ & ICM-495 \\
        Road type & \begin{tabular}[c]{@{}l@{}}1: highway, 2: arterial,\\ 3: minor road, 4: ramp,\\ 5: tunnel\end{tabular} & $L$ & Manhattan \\
        Maximum roadway \\ throughput & \begin{tabular}[c]{@{}l@{}}AADT $\times$ total lanes\end{tabular} & $MT$ & Manhattan \\
        \hline
    \end{tabular}
\end{table}

\subsection{Experimental Settings}
We compared RCDGCN against several deep learning models including both temporal and spatio-temporal architectures. The baseline models are:
\begin{itemize}
\item \textbf{FCN}(\cite{rumelhart1986learning}): A two-layer fully connected neural network.
\item \textbf{LSTM}(\cite{ma2015long}): A long-short term memory network.
\item \textbf{SGC-LSTM}(\cite{kipf2016semi}): Combines spectral graph convolution with LSTM.
\item \textbf{GAT-LSTM}(\cite{velivckovic2017graph}): Integrates a graph attention network with LSTM.
\item \textbf{TGC-LSTM}(\cite{cui2019traffic}): A traffic graph convolution neural network using LSTM with a k-hop aggregation method set to 3.
\item \textbf{ASTGCN}(\cite{guo2019attention}): Attention-based Spatial-Temporal Graph Convolution Networks.
\end{itemize}
Models were implemented using Pytorch and tested on a system equipped with an Intel Xeon CPU, an NVIDIA Titan xp GPU, and 32GB RAM. We used mean square error (MSE) for the loss function, RMSprop for optimization, and set a learning rate of 0.0001 with no weight decay. Training was conducted with a batch size of 40.

\subsection{Experiment Results}
As shown in Table \ref{predperformance}, RCDGCN outperformed all baseline models in terms of MAE and RMSE. FCN, lacking temporal and spatial dynamics, performed the worst. LSTM and its graph-enhanced versions, SGC-LSTM and GAT-LSTM, showed improved results by incorporating temporal and spatial learning mechanisms. GAT-LSTM, in particular, provided better performance due to its dynamic attention across neighboring nodes. The ASTGCN and TGC-LSTM were competitive but still lagged behind RCDGCN, which, by simulating realistic traffic propagation influenced by roadway capacity, effectively learned the complex spatio-temporal dependencies required for superior prediction accuracy.
 
\begin{table}[!ht]
    \centering
    \caption{Prediction performance in comparison to other baseline models}
    \label{predperformance}
    \small
    \begin{tabular}{|c|c|c|c|c|}
    \hline
    \multirow{2}{*}{Model} & \multicolumn{2}{c|}{ICM-495 Dataset} &
    \multicolumn{2}{c|}{Manhattan Dataset}\\
    \cline{2-5}
    & MAE & RMSE & MAE & RMSE \\
    \hline
    FCN & 5.53 & 0.64 & 4.60 & 0.62\\
    \hline
    LSTM & 3.39 & 0.42 & 3.54 & 0.50\\
    \hline
    SGC-LSTM & 3.01 & 0.39 & 3.49 & 0.49\\
    \hline
    GAT-LSTM & 2.57 & 0.29 & 3.31 & 0.46\\
    \hline
    ASTGCN & 2.51 & 0.28 & 3.06 & 0.46\\
    \hline
    TGC-LSTM & 2.50 & 0.28 & 3.01 & 0.45\\
    \hline
    \textbf{RCDGCN} & \textbf{1.80} & \textbf{0.26} & \textbf{2.55} & \textbf{0.42}\\
    \hline
    \end{tabular}
\end{table}

\section{Result Analysis}
This section investigates the effectiveness of the Traffic Attention Block (TAB) in the RCDGCN model, focusing on its impact on traffic prediction under various conditions and capacity factors. We conducted an ablation study, extracted and analyzed TAB's learned weight parameters, and executed case studies to assess model performance under traffic incident conditions.

\subsection{Ablation Experiment of TAB}
In our ablation study, we compared two versions of the model: RCDGCN with TAB and RCDGCN-R without TAB. The inclusion of TAB improved MAE and RMSE significantly, as shown in Table \ref{tab:ablation_TAB}, indicating its utility in capturing dynamic traffic conditions effectively, especially in scenarios involving traffic incidents. The ICM-495 dataset showed a notable improvement with TAB, while the static nature of the capacity factors in the Manhattan dataset yielded smaller gains.

\begin{table}[!ht]
    \centering
    \caption{Ablation experiment of external features}
    \small
    \begin{tabular}{|c|c|c|c|c|}
    \hline
    \multirow{2}{*}{Model} & \multicolumn{2}{c|}{ICM-495 Dataset} & \multicolumn{2}{c|}{Manhattan Dataset}\\
    \cline{2-5}
    & MAE & RMSE & MAE & RMSE\\
    \hline
        RCDGCN-R & 1.92 & 0.27 & 2.65 & 0.43\\
        RCDGCN & 1.80 & 0.26 & 2.59 & 0.42\\
        \hline
    \end{tabular}
    \label{tab:ablation_TAB}
\end{table}

\subsection{Graph Weight Analysis of Capacity Factors}
\begin{figure}[!tbp]
  \centering
  \subfloat[]{\includegraphics[width=0.5\columnwidth]{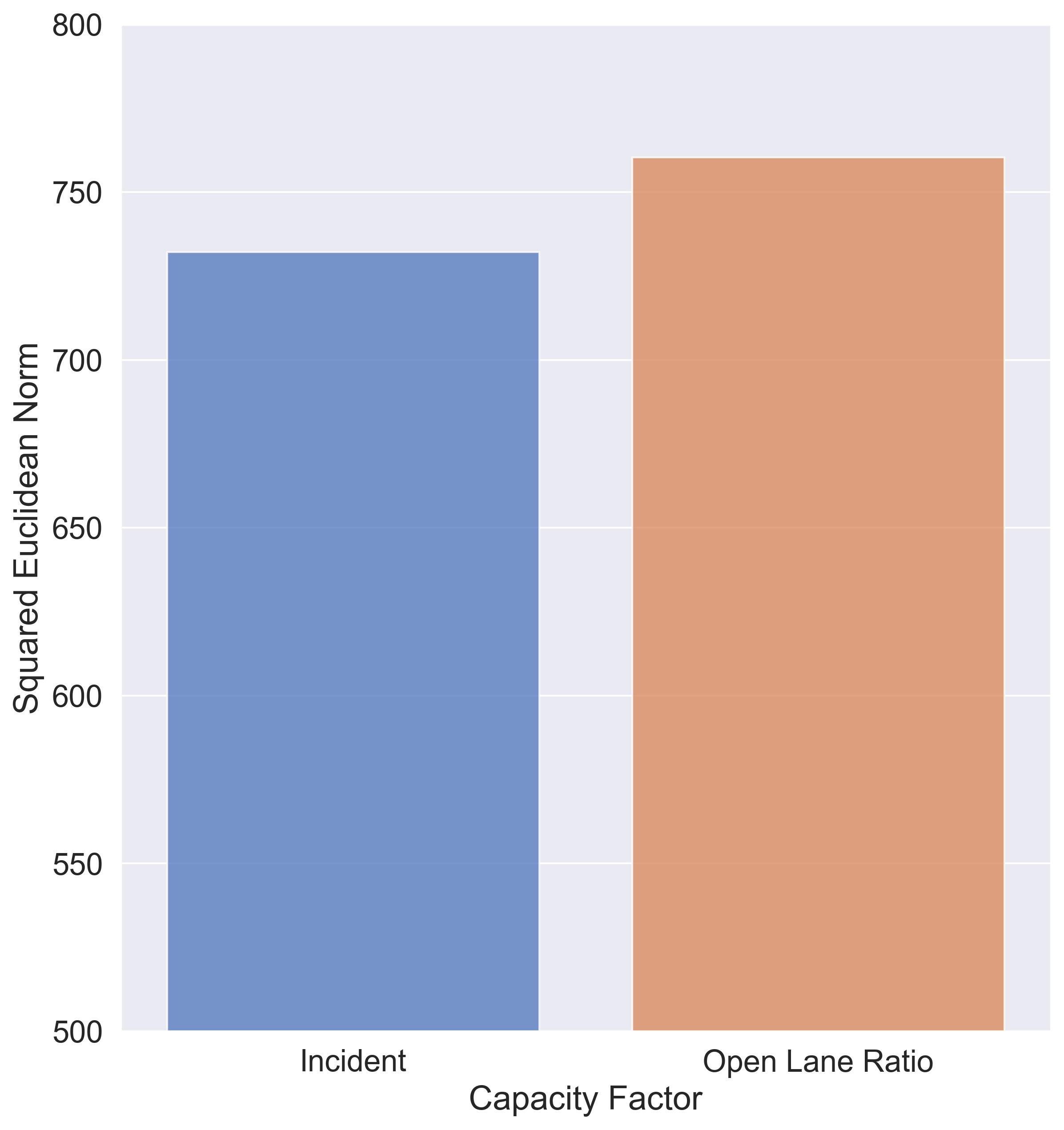}\label{fig:bar_495_matrix}}
  \hfill
  \subfloat[]{\includegraphics[width=0.5\columnwidth]{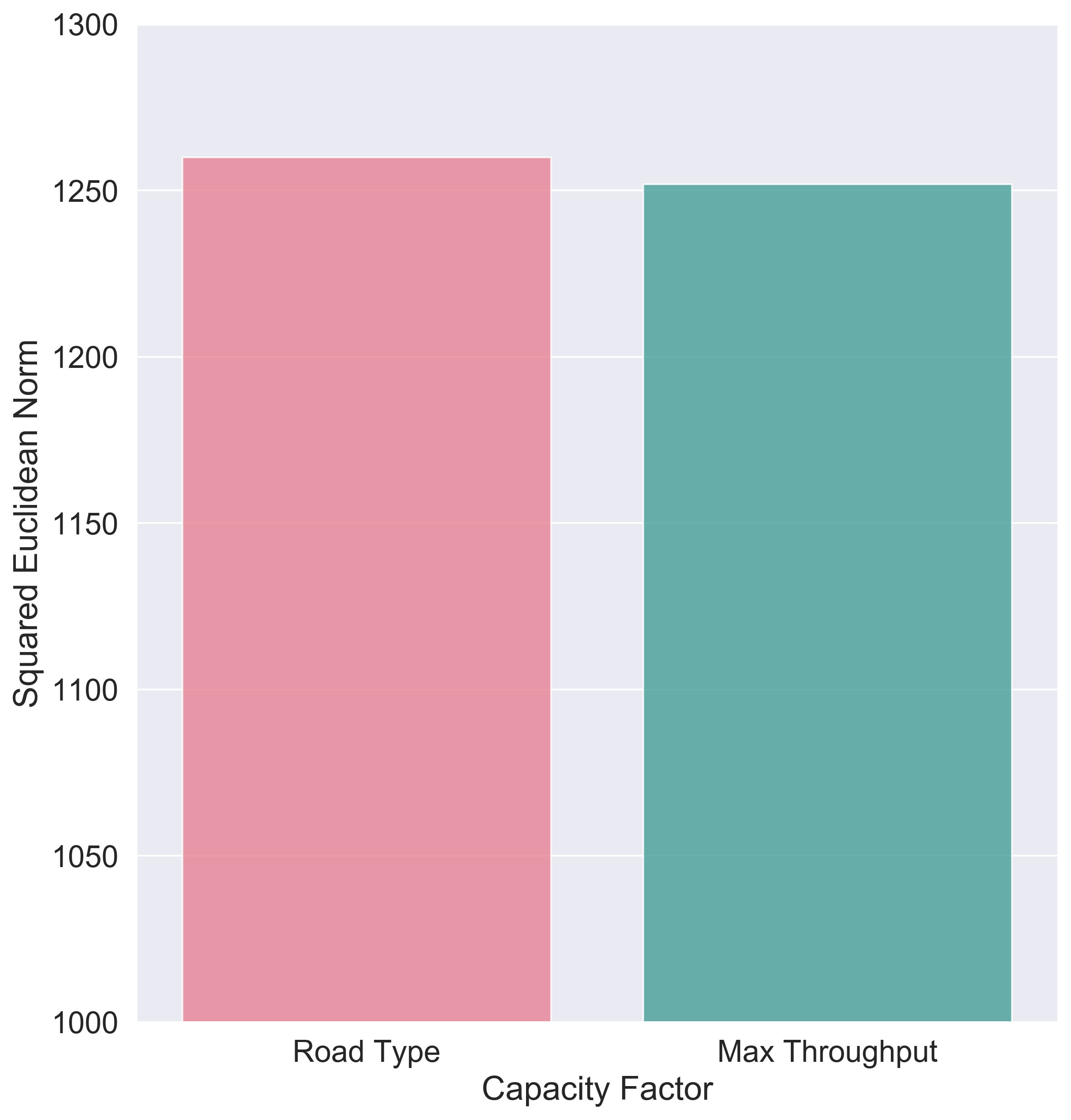}\label{fig:bar_man_matrix}}
  \caption{Matrix level squared Euclidean norms of capacity factors (a) ICM-495, (b) Manhattan}
  \label{fig:bar_matrix_twodata}
\end{figure}

\begin{figure}[htbp]
    \centering
    \includegraphics[width=1\columnwidth]{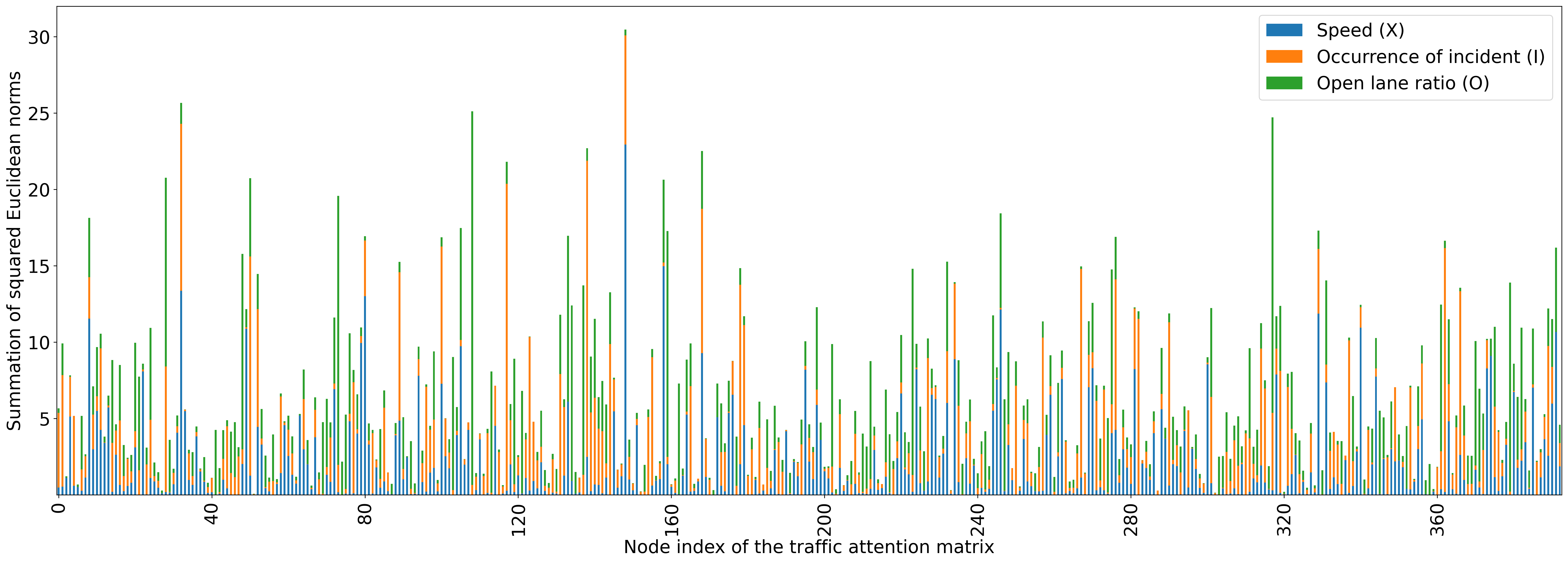}
    \caption{Bar plot of the summation of row-wise squared Euclidean norms of external factors}
    \label{fig:bar_external_link}
\end{figure}

To enhance understanding of the Traffic Attention Block (TAB), we analyzed the learned parameters of capacity factors within the traffic attention matrix $\Tilde{A}$. We assessed the impact of these factors by calculating squared Euclidean norms $\sum_{i=1}^{P} \sum_{j=1}^{Q} M_{ij}^2$, which reflect the scale of learned weights and their contributions to prediction outcomes. These norms were evaluated at both the link and matrix levels. At the matrix level, Figure \ref{fig:bar_495_matrix} demonstrates that the open lane ratio's summed squared norm surpasses that of incident occurrence, indicating a stronger impact on traffic prediction due to its direct relevance to dynamic capacity changes in road segments. Conversely, Figure \ref{fig:bar_man_matrix} illustrates a minor variance between the norms for road type and maximum throughput, suggesting similar contributions to traffic predictions by these static capacity factors. The link level is shown in Figure \ref{fig:bar_external_link}, reveals varied influences of capacity factors across individual road links. Traffic speed typically dominates, except in cases of significant incidents where factors like open lane ratio and incident occurrence become more influential, reflecting their direct impact on traffic dynamics.

\subsection{Significant Road Segment Identification}
\begin{figure*}[htp]
  \centering
  \subfloat[]{\includegraphics[width=0.728\textwidth]{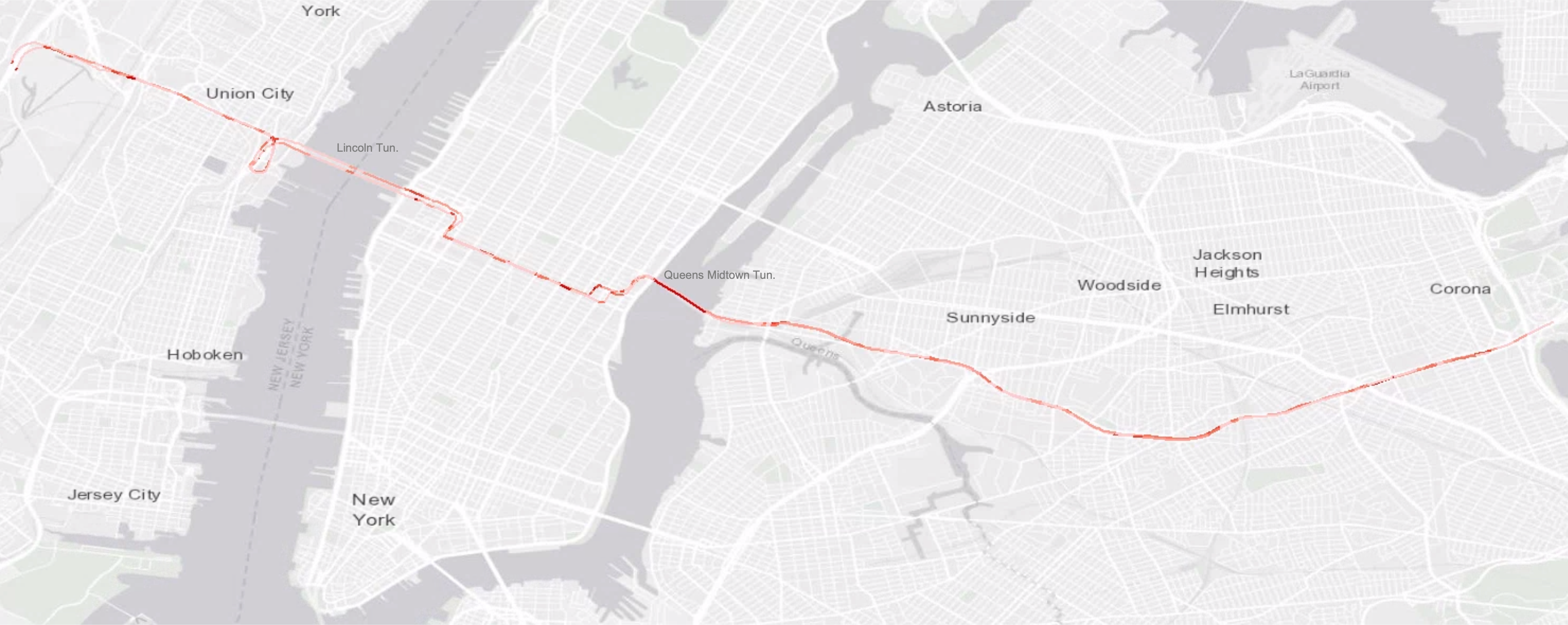}\label{fig:495hotspot}}
  \hfill
  \subfloat[]{\includegraphics[width=0.272\textwidth]{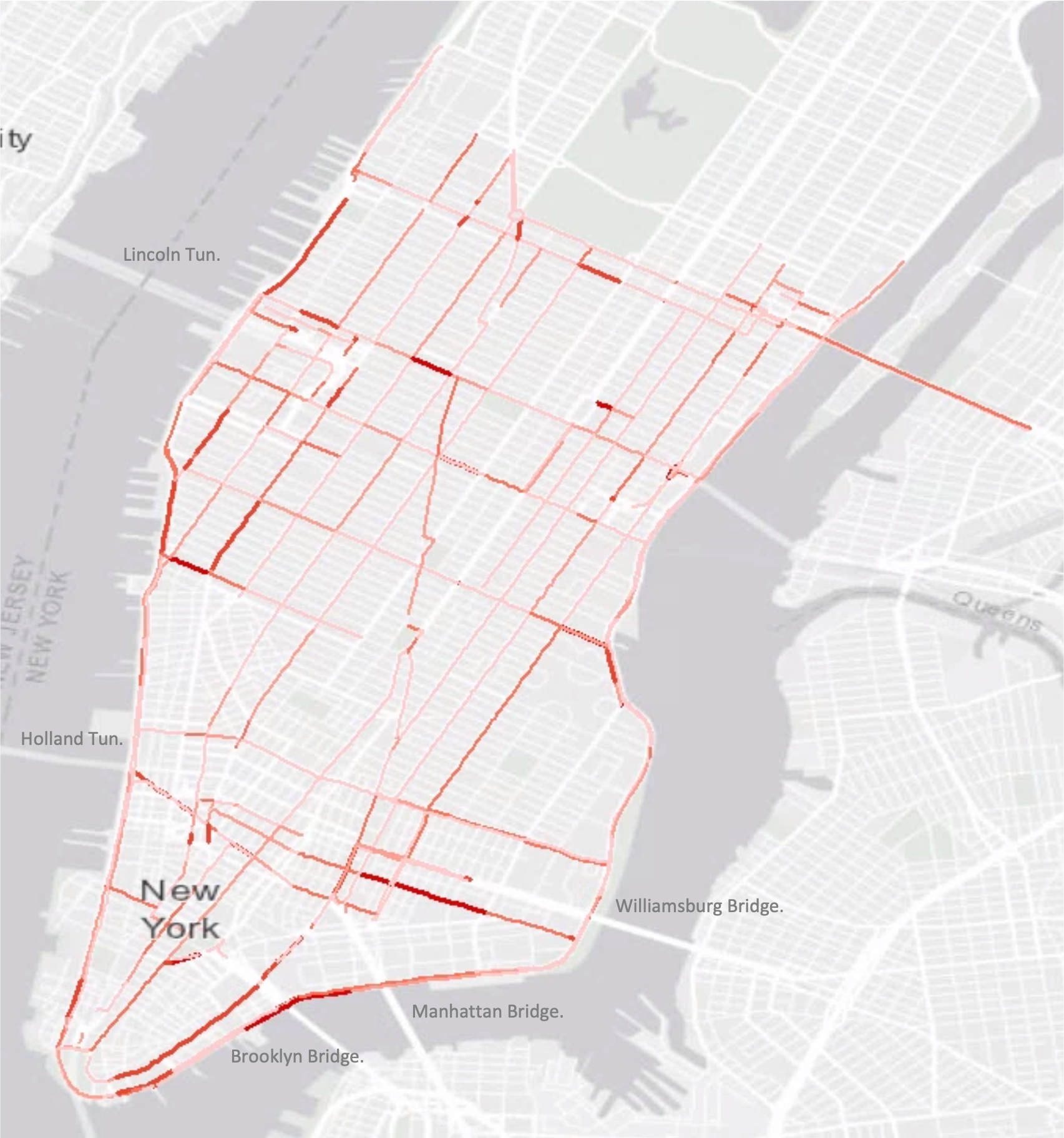}\label{fig:middownhotspot}}
  \caption{Most significant road links in ICM-495 (a), Manhattan (b)}
  \label{fig:twohotspots}
\end{figure*}

We mapped the squared Euclidean norms calculated from the learned traffic attention matrix $\Tilde{A}$ to each road link, using the row-wise elements defined as $\sum_{j=1}^N \Tilde{A}_{kj}^2$. The links whose squared norms ranked in the top 5 percentile are highlighted in dark red in Figure \ref{fig:twohotspots}. Intuitively, road links with high squared norms depict the most influential links in the transportation network.\par
To interpret these significant road segments, we investigated several empirical congestion areas. For ICM-495 in Figure \ref{fig:495hotspot}, major highlighted links were located around the Lincoln Tunnel and Queens Midtown Tunnel areas. This is reasonable since both tunnels serve as major inter-borough connections between New Jersey, Manhattan, and Queens, carrying intra- and inter-borough traffic. For the Manhattan network in Figure \ref{fig:middownhotspot}, highlighted links focused on urban areas around the Lincoln Tunnel, Brooklyn Bridge, and Manhattan Bridge, known congested intra-borough routes between Manhattan and Brooklyn. Other highlighted urban streets, like Grand Street in Lower Manhattan, are located between major busy areas like SOHO and Chinatown, which typically carry heavy daily traffic. In summary, the learned traffic attention matrix helps interpret model results for transportation networks containing inter-borough and intra-borough traffic by identifying significant traffic hotspot areas.

\subsection{Case Studies}
\begin{figure}[!tbp]
  \centering
  \subfloat[Case 1: Accident from 14:02 to 15:10, 02/25, right lane  blocked on link ID 24892517 ]{\includegraphics[width=0.8\linewidth]{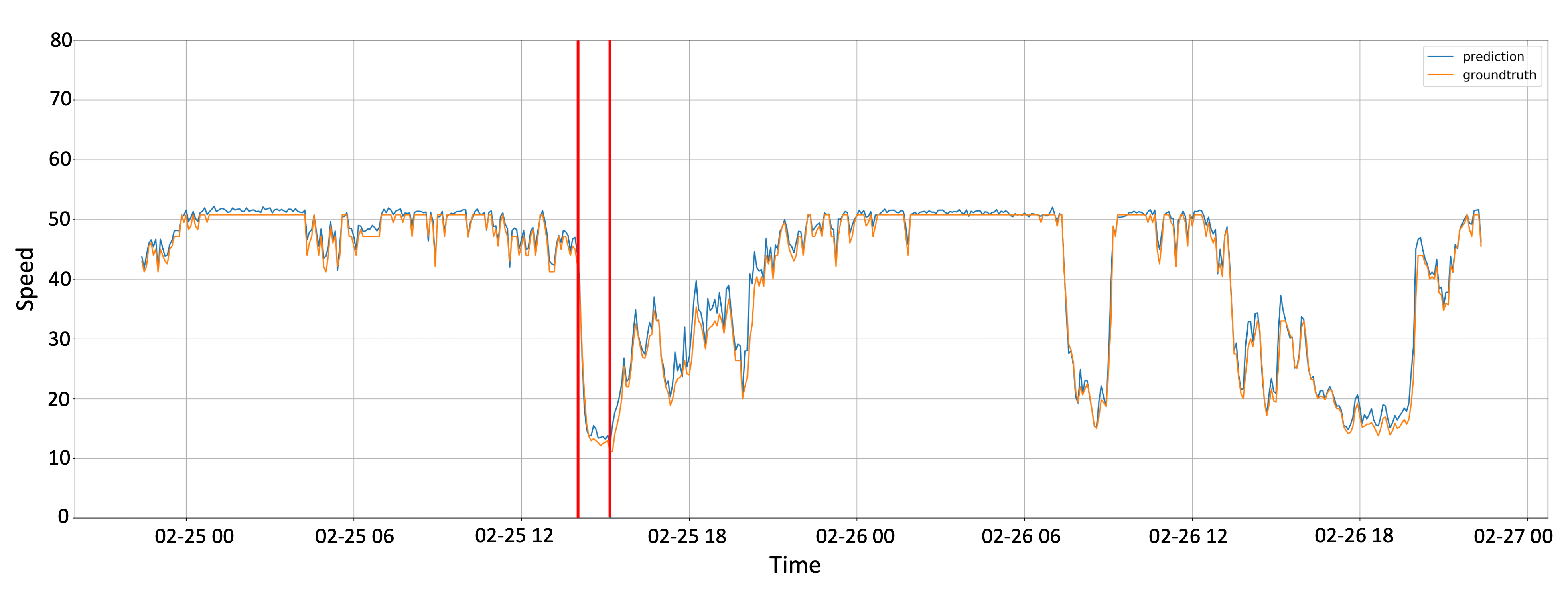}\label{fig:case1_acc}}
  \hfill
  \subfloat[Case 2: Emergency construction from 13:11 to 16:12, 02/26, left lane closed on link ID 858823464]{\includegraphics[width=0.8\linewidth]{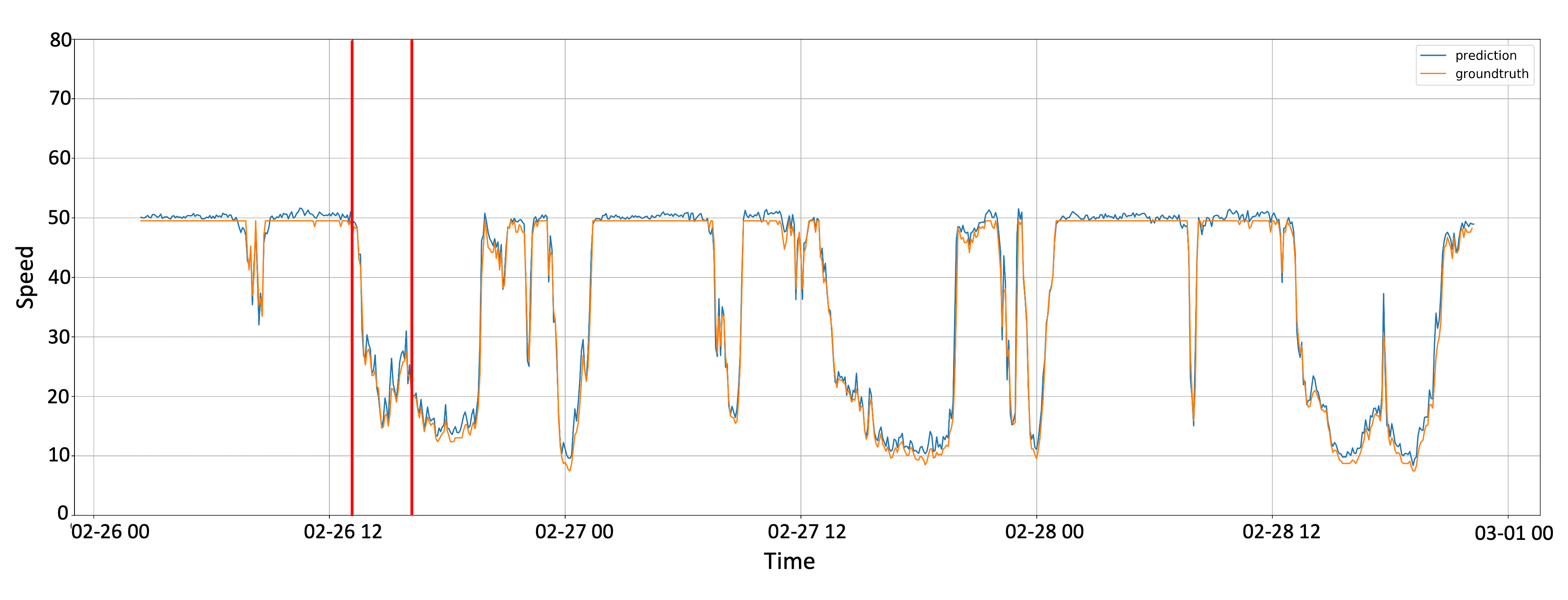}\label{fig:case2_emer}}
  \hfill
  \subfloat[Case 3: Construction from 21:43, 02/23 to 14:39, 02/25, all lanes closed on link ID 733067100]{\includegraphics[width=0.8\linewidth]{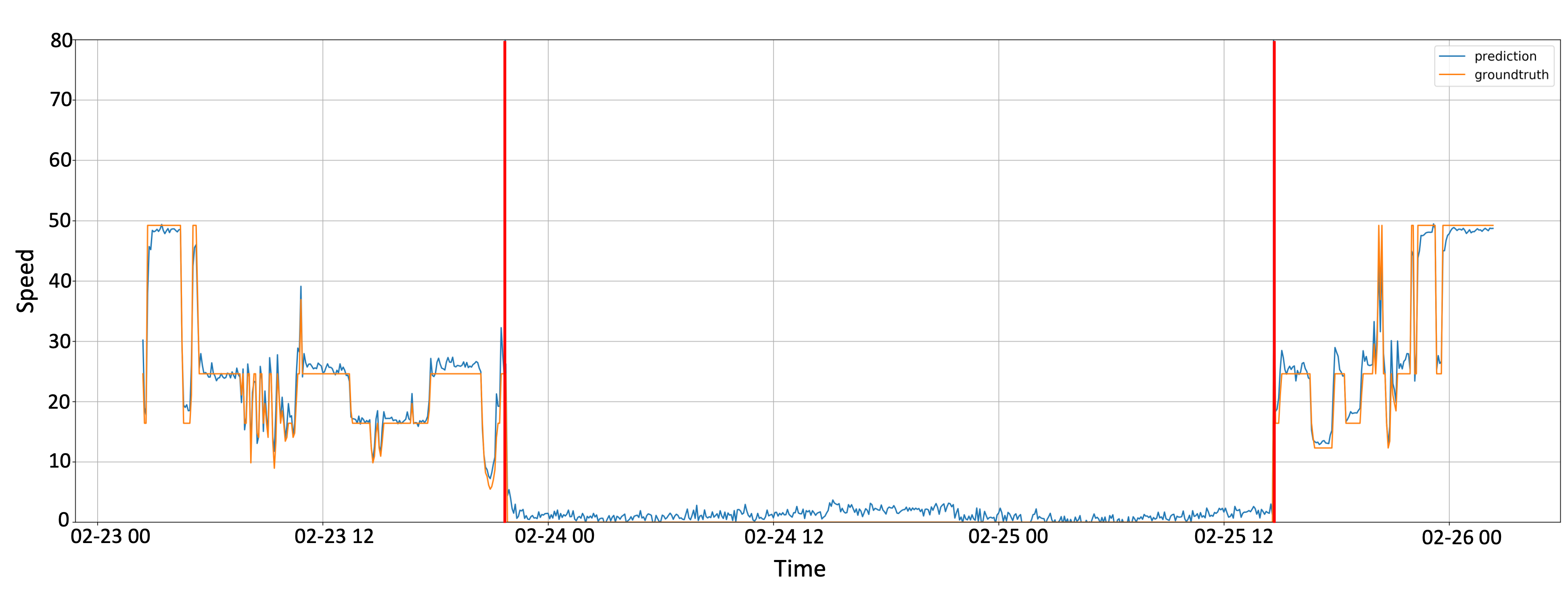}\label{fig:case3_construc}}
  \caption{Case studies with different traffic incidents}
  \label{fig:case_study}
\end{figure}

We compared predicted and ground truth traffic speeds for selected ICM-495 road segments, focusing on cases with traffic incidents to validate RCDGCN's ability to capture resulting interruptions. We examined three incident types: accident (sudden disruption with partial lane closure), emergency construction (unplanned event with partial closure), and pre-planned construction (planned all-lane closure). Overall, as shown in Figure \ref{fig:case_study}, predicted speeds fit well with ground truth data.\par 
Case 1 (Figure \ref{fig:case1_acc}) showed an accident around 2-3 PM on Sunday, blocking the right lane and causing a sudden speed drop. The blue vertical lines indicate accident start/end times, and RCDGCN captured the sudden drop and recovery trend after clearance.
Case 2 (Figure \ref{fig:case2_emer}) involved an emergency construction with a left lane closure lasting about 3 hours on Monday afternoon. The speed trend differed from the next day due to the event. RCDGCN captured this variance with the traffic incident information, showing speed recovery approaching construction end and returning to Tuesday's trend after removal.
In contrast to partial closures, Case 3 (Figure \ref{fig:case3_construc}) had a pre-planned all-lane closure construction event, causing the strongest traffic interruption with zero ground truth speeds. RCDGCN managed to capture this heavy interruption with a flat trend of small speed values. In summary, we investigated cases with special traffic incidents and showed RCDGCN can capture sudden disruptions under different interruption levels.

\section{Conclusion} \label{sc:conclusion}


In this study, we developed the Roadway Capacity Driven Graph Convolution Network (RCDGCN) to enhance traffic state predictions by integrating external capacity factors. Unlike traditional GNN-based models, RCDGCN employs a Traffic Attention Block (TAB) that effectively adapts to varying traffic conditions by assigning intensity scores to interactions between road links. This model outperformed leading deep learning methods in prediction accuracy, as evidenced by an ablation experiment and graph weight analysis, which confirmed the substantial impact of different capacity factors on prediction outcomes. Additionally, the traffic attention matrix from TAB proved instrumental in identifying critical road links, further validated through empirical analyses and case studies that demonstrated RCDGCN's ability to detect significant disruptions from traffic incidents.

\newpage
\section{Acknowledgements}
This work was supported by the C2SMARTER, a Tier 1 U.S. Department of Transportation (USDOT) funded University Transportation Center (UTC) led by New York University funded by USDOT. The contents of this paper only reflect the views of the authors who are responsible for the facts and do not represent any official views of any sponsoring organizations or agencies.





\bibliographystyle{IEEEtran}

\bibliography{reference}

\end{document}